\documentclass[11pt]{article}
\usepackage{coling2020}
\usepackage{comment}
\usepackage{times}
\usepackage{url}
\usepackage{latexsym}
\usepackage{color}
\usepackage[dvipsnames]{xcolor}

\usepackage{booktabs}
\usepackage{amsmath,amssymb}
\usepackage{booktabs}
\usepackage{graphicx}
\usepackage{multirow}

\colingfinalcopy 

\title{Don't Patronize Me! An Annotated Dataset with Patronizing and Condescending Language towards Vulnerable Communities}

\author{Carla Pérez-Almendros \hspace{0.8cm} Luis Espinosa-Anke \hspace{0.8cm} Steven Schockaert \\
         School of Computer Science and Informatics \\
	    Cardiff University, United Kingdom\\
	     {\tt \{perezalmendrosc,espinosa-ankel,schockaerts1\}@cardiff.ac.uk}}

\date{}

\begin{document}
\maketitle
\begin{abstract}
In this paper, we introduce a new annotated dataset which is aimed at supporting the development of NLP models to identify and categorize language that is patronizing or condescending  towards vulnerable communities (e.g.\ refugees, homeless people, poor families). While the prevalence of such language in the general media has long been shown to have harmful effects, it differs from other types of harmful language, in that it is generally used unconsciously and with good intentions. We furthermore believe that the often subtle nature of patronizing and condescending language (PCL) presents an interesting technical challenge for the NLP community. Our analysis of the proposed dataset shows that identifying PCL is hard for standard NLP models, with language models such as BERT achieving the best results. 
\end{abstract}

\section{Introduction}

\blfootnote{
  %
  %
  %
  %
  \hspace{-0.65cm} 
  This work is licensed under a Creative Commons 
  Attribution 4.0 International Licence.
  Licence details:
  \url{http://creativecommons.org/licenses/by/4.0/}.
  %
  %
  }


In this paper, we analyze the use of Patronizing and Condescendig Language (PCL) towards vulnerable communities in the media. An entity engages in PCL
when its language use shows a superior attitude towards others or depicts them in a compassionate way. This effect is not always conscious and the intention of the author is often to help 
the person or group they refer to (e.g.\ by raising awareness or funds, or moving the audience to action). However, these superior attitudes and a discourse of pity 
can routinize discrimination and make it less visible \cite{ng2007language}. Moreover, general media publications reach a large audience and we believe that unfair treatment of vulnerable groups in such media might lead to greater exclusion and inequalities.


While there has been substantial work on modelling language that purposefully undermines others, e.g.\ offensive language or hate speech \cite{zampieri2019semeval,basile2019semeval}, the modelling of PCL is still an emergent area of study in NLP. Some reasons for this might include that the use of PCL in the media is commonly unconscious, subtler and more subjective than the types of discourse that are typically targeted in NLP. Specifically, a special focus in PCL towards vulnerable communities has not been yet considered, to the best of our knowledge.

Within a broader setting, there has been some work on PCL which is concerned with the communication between two parties, where one is patronized by the other, 
such as in social media interactions. In particular, \newcite{wang2019talkdown} recently published the Talkdown corpus for condescension detection in comment-reply pairs from Reddit. In this work, the authors highlight the difficulty of the task and the need for a high-quality dataset annotated by experts, which is the approach we take for studying PCL towards vulnerable communities. 

To encourage more research on detecting PCL language, we introduce the \textit{Don't Patronize Me! dataset}\footnote{Available at \url{https://github.com/Perez-AlmendrosC/dontpatronizeme}.}. This dataset contains more than 10,000 paragraphs extracted from news stories, which have been annotated to indicate the presence of PCL at the text span level. The paragraphs were selected to cover English language news sources from 20 different countries, covering different types of vulnerable communities (e.g.\ homeless people, immigrants and poor families). We furthermore propose a taxonomy of PCL categories, focused on PCL towards vulnerable communities. Each of the PCL text spans from our dataset has been annotated with a category label from this taxonomy. Finally, we also provide some analysis of the dataset. Among others, we find that even simple baselines are able to detect PCL to some extent, which suggests that this task is feasible for NLP systems, despite the subtle nature of PCL. On the other hand, we also find that the considered models, including approaches based on BERT \cite{DBLP:conf/naacl/DevlinCLT19}, struggle to detect certain categories of PCL, suggesting that there is still considerable room for improvement. In particular, while some forms of PCL can be detected by identifying relatively simple linguistic patterns, many other cases seem to require a non-trivial amount of world knowledge.





%
%

\section{Related Work}
Condescending and patronizing treatment has been widely studied in various fields, such as language studies \cite{margic2017communication}, sociolinguistics \cite{giles1993patronizingelderly}, politics \cite{huckin2002critical} or medicine \cite{komrad1983medicalpaternalism}. Within NLP, there has been extensive work on several forms of harmful language, but this work has generally focused on explicit, aggressive and flagrant phenomena such as fake news detection \cite{conroy2015automatic}; trust-worthiness prediction and fact-checking \cite{atanasova2018overview,atanasova2019overview}; modeling offensive language, both generic \cite{zampieri2019semeval}, and geared towards specific communities \cite{basile2019semeval}; or rumour propagation \cite{derczynski2017semeval}. 
Recently, however, some work on condescending language has started to appear. For instance, \newcite{wang2019talkdown} 
introduced the task of modelling condescension in direct communication from an NLP perspective, and developed a dataset with annotated social media messages. In the same year, \newcite{sap2019powerimplications} discussed the social and power implications behind certain uses of language, an important concept in the unbalanced power relations that are often present in condescending treatment. Also related to unfair treatment of underprivileged groups, \newcite{mendelsohn2020dehumanization} analyzed, from a computational linguistics point of view, how language has dehumanized minorities in news media over time. 




\section{Background on PCL}\label{secTraitsPCL}
Research in sociolinguistics has suggested the following traits of PCL towards vulnerable communities: 
\begin{itemize}
  \item it fuels discriminatory behaviour by relying on subtle language \cite{mendelsohn2020dehumanization};\vspace{-0.2cm} 
  \item it creates and feeds stereotypes \cite{fiske1993stereotyping}, which drive to greater exclusion, discrimination, rumour spreading and misinformation \cite{nolan2013things};\vspace{-0.2cm} 
  \item it strengthens power-knowledge relationships \cite{foucault1980power}, positioning one community as superior to others;\vspace{-0.2cm}
  \item it usually calls for charitable action instead of cooperation, so communities in need are presented as passive receivers of help, unable to solve their own problems and waiting for a \textit{saviour} to help them out of their situation \cite{bell2013raising,straubhaar2015stark};\vspace{-0.2cm}
  \item it tends to avoid stating the reasons for very deep-rooted societal problems, by concealing those responsible or even, in some cases, by apportioning blame to the underprivileged communities or individuals themselves; \vspace{-0.2cm}
  \item it proposes ephemeral and simple solutions \cite{chouliaraki2010post-humanitarian}, which oversimplify the wicked problems \cite{head2008wicked} vulnerable communities face. 
\end{itemize}

\noindent The use of PCL makes it more difficult for vulnerable communities to overcome difficulties and reach total inclusion \cite{nolan2013things}.



\subsection[*]{How to identify PCL?}

In this work, we analyze discourse on vulnerable communities. We will consider a piece of text as containing PCL when, referring to an underprivileged individual or community, we can identify one or several of the following traits:
\begin{itemize}
  \item The use of the language states the differences between the \textit{`us'} and the \textit{`them'}. The vulnerable community is depicted as different to \textit{us}, with other experiences and life stories. This discourse establishes an invisible distance between the two communities.
  
  \item The language raises a feeling of pity towards the vulnerable community, for example by using (or abusing) adjectives or by recurring to flowery words to depict a certain situation in a literary way (i.e., metaphors, euphemisms or hyperboles).
  
  \item The author and the community they belong to are presented as \textit{saviours} of those in need. Not only do they have the capacity to solve their problems, but also a moral responsibility to do so. The superior or privileged community is also presented as having the knowledge and experience to face and solve the problems of the vulnerable ones.
  
  \item In the opposite direction, the members of the vulnerable community are described as lacking the privileges the author's community enjoys, or even the knowledge or experience to overcome their own problems. They will need, therefore, the help of others to improve their situation. 
  \item The vulnerable community and its members are presented either as victims (i.e.\ overwhelmed, victimized or pitied) or as heroes just because of the situation they face.
\end{itemize}

\subsection{What is not PCL?}

Precisely because we are studying the discourse towards vulnerable communities, it can be easy to classify a piece of text as condescending mistakenly. We want to highlight, in particular, the following 
two situations where the language that is used to talk about unprivileged groups is not condescending. 

\begin{itemize}
  \item Because they are experiencing vulnerability, the news about them often depicts rough situations. The description of an extreme situation can be harsh and stark and leave the reader with a feeling of sadness and helplessness, while not necessarily being condescending.  

  \item With PCL, the superiority of the author is concealed behind a friendly or compassionate approach towards the situation of vulnerable communities. Thus, a message which is openly offensive, aggressive or containing prejudiced, discriminatory or hate speech is not considered to be PCL for the purpose of our dataset.
  
\end{itemize}

\section{The Don't Patronize Me! dataset}
The Don't Patronize Me! dataset currently contains 10,637 paragraphs about potentially vulnerable social groups. These paragraphs have been selected from general news stories and have been annotated with labels that indicate the type of PCL language that is present, if any.
The paragraphs have been extracted from the News on Web (NoW) corpus\footnote{The corpus is used with the permission of its author.} \cite{nowcorpus}. To this end, we first selected ten keywords related to potentially vulnerable communities widely covered in the media and susceptible of receiving a condescending or patronizing treatment: 
\textit{disabled}, \textit{homeless}, \textit{hopeless}, \textit{immigrant}, \textit{in need}, \textit{migrant}, \textit{poor families}, \textit{refugee}, \textit{vulnerable} and \textit{women}. 
Next, we retrieved paragraphs in which these keywords are mentioned, choosing a similar number of paragraphs for each of the 10 keywords and each of the 20 English speaking countries that are covered in the NoW corpus. An overview of the number of paragraphs for each keyword-country combination can be found in Table \ref{tab:CountryKwStatistics}.
All the selected paragraphs come from news stories that were published between 2010 and 2018. The data was annotated by three expert annotators, with backgrounds in communication, media and data science. Two annotators annotated the whole dataset (\textit{ann1} and \textit{ann2}), while the third one (\textit{ann3}) acted as a referee to provide a final label in case of disagreements. An extended data statement \cite{bender2018data} about the corpus will be published together with the dataset.

\begin{table}[]
\centering
\footnotesize
\begin{tabular}{lccccccccccrc}
\toprule
               & dis  & hom  & hop  & imm  & need & mig  & poor & ref  & vul  & wom  & \textbf{Total} \\ \midrule
Australia      & 56   & 51   & 52   & 56   & 57   & 57   & 54   & 54   & 60   & 55   & 552            \\
Bangladesh     & 51   & 57   & 46   & 50   & 51   & 56   & 46   & 52   & 55   & 53   & 517            \\
Canada         & 53   & 53   & 52   & 51   & 52   & 47   & 55   & 56   & 61   & 52   & 532            \\
Ghana          & 62   & 55   & 57   & 56   & 51   & 58   & 25   & 53   & 54   & 55   & 526            \\
Hong Kong      & 60   & 58   & 32   & 53   & 55   & 59   & 22   & 49   & 52   & 61   & 501            \\
Ireland        & 61   & 49   & 55   & 58   & 58   & 58   & 36   & 58   & 48   & 55   & 536            \\
India          & 53   & 52   & 62   & 60   & 57   & 52   & 52   & 58   & 59   & 50   & 555            \\
Jamaica        & 53   & 62   & 47   & 56   & 58   & 51   & 11   & 54   & 50   & 51   & 493            \\
Kenya          & 52   & 51   & 55   & 56   & 51   & 54   & 55   & 49   & 57   & 61   & 541            \\
Sri Lanka      & 53   & 57   & 57   & 59   & 48   & 53   & 32   & 56   & 49   & 50   & 514            \\
Malaysia       & 58   & 48   & 47   & 54   & 62   & 58   & 53   & 58   & 60   & 56   & 554            \\
Nigeria        & 55   & 60   & 49   & 52   & 53   & 56   & 49   & 56   & 60   & 55   & 545            \\
New Zealand    & 63   & 49   & 61   & 51   & 50   & 56   & 51   & 49   & 49   & 47   & 526            \\
Philipines     & 61   & 56   & 56   & 48   & 54   & 59   & 53   & 51   & 55   & 52   & 545            \\
Pakistan       & 50   & 55   & 51   & 51   & 58   & 57   & 57   & 56   & 54   & 56   & 545            \\
Singapore      & 51   & 56   & 53   & 57   & 59   & 59   & 54   & 45   & 54   & 50   & 538            \\
Tanzania       & 51   & 55   & 18   & 53   & 50   & 51   & 38   & 48   & 53   & 51   & 468            \\
UK             & 55   & 50   & 47   & 55   & 56   & 53   & 59   & 58   & 58   & 51   & 542            \\
United States  & 58   & 60   & 54   & 51   & 54   & 55   & 53   & 61   & 47   & 58   & 551            \\
South Africa   & 60   & 54   & 63   & 58   & 55   & 54   & 59   & 50   & 47   & 56   & 556            \\ \midrule
\textbf{Total} & 1116 & 1088 & 1014 & 1085 & 1089 & 1103 & 914  & 1071 & 1082 & 1075 & 10637          \\ \bottomrule
\end{tabular}%
\caption{Number of paragraphs per keyword and country in the dataset.  The considered keywords are disabled (dis), homeless (hom), hopeless (hop), immigrant (imm), in-need (need), migrant (mig), poor-families (poor), refugees (ref), vulnerable (vul) and women (wom).}
\label{tab:CountryKwStatistics}
\end{table}

\subsection{Categories of PCL towards vulnerable communities}
For all text spans that were annotated as containing PCL, the annotators also provided a category label. This allows us to analyze at a finer-grained level to what extent NLP models are able to recognize the different traits of PCL. These labels might also make it easier to train NLP models for detecting PCL, for instance by treating them as privileged information during training \cite{vapnik2009new}. Inspired by the characteristics of PCL discussed in Section \ref{secTraitsPCL}, we have used the following seven categories, which we grouped into three higher-level categories. 
    \begin{itemize}

    \item \textbf{The saviour.} The community which the author and the majority of the audience belong to is presented in some way as \textit{saviours} of those vulnerable or in need. The language used subtly positions the author in a better, more privileged situation than the vulnerable community. They express the will to help them, from their superior and advantageous position. There is a clear difference between the \textit{we} and the \textit{they}. As part of this category, we can find examples of the following subcategories:
  
  \begin{itemize}
  \item \textbf{ Unbalanced power relations.} By means of the language, the author distances themselves from the community or the situation they are talking about, and expresses the will, capacity or responsibility to help them. It is also present when the author entitles themselves to give something positive to others in a more vulnerable situation, especially when what the author \textit{concedes} is a right which they do not have any authority to decide to give. 
  \begin{quote}
    (i.e.\ \textit{`You can make a difference in their lives'} or \textit{`They come back in with nothing and we need to outfit them again'} or \textit{`They deserve another opportunity'} or \textit{`They also have the right to love'}).
  \end{quote}
  
  \item \textbf{Shallow solution.} A simple and superficial charitable action by the privileged community is presented either as life-saving/life-changing for the unprivileged one, or as a solution for a deep-rooted problem. 
  \begin{quote}
    (i.e.\ \textit{`Raise money to combat homelessness by curling up in sleeping bags for one night'} or \textit{`If every supporter on Facebook donated just one box each it would make a real difference to many poor families'}).
  \end{quote}
  \end{itemize}
  \item \textbf{The expert.} The underlying message is that the privileged community, which the author and their audience belong to, knows better what the vulnerable community needs, how they are or what they should do to overcome their situation. We consider the following subcategories:
 \begin{itemize}
 
 \item \textbf{ Presupposition}, when the author assumes a situation as certain without having all the information, or generalises their or somebody else's experience as a categorical truth without presenting a valid, trustworthy source for it (e.g.\ a research work or survey). The use of stereotypes or clich\'es are also considered to be examples of presupposition.
 
 \begin{quote}
    (i.e.\ \textit{`[...] elderly or disabled people who are simply unable to evacuate due to physical limitations'} or \textit{`If the economy fills with women, it will develop beautifully'});
  \end{quote}
  
  \item \textbf{ Authority voice}, when the author stands themselves as a spokesperson of the group, or explains or advises the members of a community about the community itself or a specific situation they are living.
  \begin{quote}
    (i.e.\ \textit{`Accepting their situation is the first step to having a normal life'} or \textit{`We also know that they can benefit by receiving counseling from someone who can help them understand.'});
  \end{quote}
  
  \end{itemize}
  \item \textbf{The poet.} The focus is not on the \textit{we} (author and audience), but on the \textit{they} (the individual or community referred to). The author uses a literary style to describe people or situations. They might, for example, use (or abuse) adjectives or rhetorical devices to either present a difficult situation as somehow beautiful, something to admire and learn from, or they might carefully detail its roughness to touch the heart of their audience. The subcategories we establish are:
  \begin{itemize}
  \item \textbf{Metaphor}. They can conceal PCL, as they cast an idea in another light, making a comparison between unrelated concepts, often with the objective of depicting a certain situation in a softer way. For the annotation of this dataset, euphemisms are considered as an example of metaphors. 
  \begin{quote}
    (i.e.\ \textit{`Poor children might find more obstacles in their race to a worthy future'} or \textit{`those who cling to boats to reach a shore of survival'});
  \end{quote}
  \item \textbf{Compassion}. The author presents the vulnerable individual or community as needy,
  raising a feeling of pity and compassion from the audience towards them. It is commonly characterized by the use of flowery wording that does not provide information, but the author enjoys the detailed and poetic description of the vulnerability; 
  \begin{quote}
    (i.e. \textit{`Some are lured by corrupt ``agents", smuggled across the searing Sahara and discarded in the streets of Europe, resigned to selling fake designer bags as undocumented immigrants'} or \textit{`For the roughly 2,000 migrants who call it home, the broken windows and decaying walls of the decrepit warehouse offer scant respite from the harsh blizzard conditions currently striking Serbia'}).
  \end{quote}
  \item \textbf{The poorer, the merrier}. The text is focused on the community, especially on how the vulnerability makes them better (e.g. stronger, happier or more resilient) or how they share a positive attribute just for being part of a vulnerable community. People living vulnerable situations have values to admire and learn from. The message expresses the idea of vulnerability as something beautiful or poetic. We can think of the typical example of `poor people are happier because they don't have material goods'. 
  \begin{quote}
  
    (i.e.\ \textit{`He is reminded of the true meaning of hope by people living in situations the world would see as hopeless'} or \textit{`her mom is disabled and living with her gives her strength to face everyday's life'} or
    \textit{`refugees are wonderful people'})
  \end{quote}
  \end{itemize} 
  \end{itemize}
Finally, in the dataset, we also included an ``Other'' category, to classify all the text spans which the annotators considered to contain PCL, but which they could not assign to any of the previous categories. However, the annotators did not need to use this label for any instance.

\subsection{Annotation}

To annotate the dataset, a two-step process has been followed. In the first step, annotators determined which paragraphs contain PCL. Subsequently, in the second step, the annotators indicated which text spans within these paragraphs contain PCL and they labelled each of these text spans with a particular PCL category. We now discuss these two steps in more detail.

\subsubsection{Step 1: Paragraph-Level Identification of PCL} The aim of this annotation step is to decide for each paragraph whether or not it contains PCL.
%
This annotation step proved more difficult than expected, stemming from the often subtle and subjective nature of PCL. To mitigate this, we decided to annotate the paragraphs with three possible labels: 0, meaning that the paragraph does not contain PCL, 1, meaning that it is considered to be a borderline case, or 2, meaning that it clearly contains PCL. 
We computed the Kappa Inter-Annotator Agreement (IAA) between two main annotators (\textit{ann1} and \textit{ann2}) across the three labels, obtaining a moderate agreement of 41\%. 
%
If we omit all paragraphs which were marked as borderline by at least one annotator, the IAA reaches a substantial 61\% \cite{landis1977measurement}. 

Overall, ann1 and ann2 agreed in 9,182 paragraphs and disagreed in 1457. Among the disagreements, 590 were total disagreements (0 vs 2) and 867 cases included borderline cases.
To maximize the amount of information captured by the annotations, and in particular obtain a finer-grained assessment about borderline cases, we combined the labels provided by the two annotators into a 5-point scale, as follows:
  \begin{itemize}
    \item Label 0: both annotators assigned the label 0 (0 + 0).
    \item Label 1: one annotator assigned the label 0 and the other assigned the label 1 (0 + 1).
    \item Label 2: both annotators assigned the label 1 (1 + 1).
    \item Label 3: one annotator assigned the label 2 and the other assigned the label 1 (2 + 1).
    \item Label 4: both annotators assigned the label 2 (2 + 2).
  \end{itemize}

  \noindent Note how partial disagreement between the annotators is thus reflected in the final label.
  The cases of total disagreement, where one annotator labeled the instance as clearly not containing PCL and the other annotated it as clearly containing PCL (0 + 2), were annotated by \textit{ann3}. After this supplementary annotation, the paragraph is either labelled as 1, if the third annotator considered the paragraph not to contain PCL, as 2, if they considered it to be a borderline case, or as 3, if they considered the paragraph to clearly contain PCL. In this way, the labels 0 and 4 remain reserved for clear-cut cases.
  For the experimental analysis presented in this paper, we treated paragraphs with final labels 0 and 1 as negative examples (i.e.\ as instances not containing PCL) and paragraphs with final labels 2, 3 and 4 as positive examples (i.e.\ as instances containing PCL). In total, interpreted in this way, the dataset contains 995 positive examples of PCL.

\begin{table}[]
\centering
\footnotesize
\begin{tabular}{lcccccccrc}
\toprule
                       & \multicolumn{1}{l}{\textbf{unb}} & \multicolumn{1}{l}{\textbf{com}} & \multicolumn{1}{l}{\textbf{pre}} & \multicolumn{1}{l}{\textbf{aut}} & \multicolumn{1}{l}{\textbf{sha}} & \multicolumn{1}{l}{\textbf{met}} & \multicolumn{1}{l}{\textbf{merr}} & \multicolumn{1}{l}{\textbf{Total}} \\ \midrule
\textbf{Disabled}      & 96                               & 55                               & 26                               & 23                               & 21                               & 17                               & 12                                & 250                                \\
\textbf{Homeless}      & 231                              & 154                              & 38                               & 31                               & 84                               & 56                               & 6                                 & 600                                \\
\textbf{Hopeless}      & 105                              & 224                              & 95                               & 60                               & 6                                & 59                               & 6                                 & 555                                \\
\textbf{Immigrant}     & 29                               & 32                               & 21                               & 7                                & 4                                & 5                                & 4                                 & 102                                \\
\textbf{In-need}       & 347                              & 85                               & 17                               & 42                               & 85                               & 36                               & 6                                 & 618                                \\
\textbf{Migrant}       & 40                               & 45                               & 9                                & 14                               & 4                                & 10                               & 4                                 & 126                                \\
\textbf{Poor-families} & 185                              & 131                              & 63                               & 59                               & 41                               & 67                               & 11                                & 557                                \\
\textbf{Refugee}       & 93                               & 78                               & 22                               & 17                               & 33                               & 20                               & 5                                 & 268                                \\
\textbf{Vulnerable}    & 130                              & 54                               & 22                               & 41                               & 12                               & 36                               & 1                                 & 296                                \\
\textbf{Women}         & 51                               & 30                               & 34                               & 33                               & 12                               & 13                               & 9                                 & 182                                \\ \midrule
\textbf{Total}         & 1307                             & 888                              & 347                              & 327                              & 302                              & 319                              & 64                                & 3554                               \\ \bottomrule
\end{tabular}%
\caption{Number of text spans that have been labelled with each of the PCL categories, per keyword. The considered categories are unbalanced power relations (unb), compassion (comp), presupposition(pres), authority voice (auth), shallow solution (shal), metaphor (met),  and the poorer, the merrier (merr).}
\label{tab:kw-category}
\end{table}
  

\subsubsection{Step 2: Identifying Span-Level PCL Categories} Those paragraphs labelled as containing PCL in Step 1 are collected for further annotation. The aim of this second step is to specify which text spans within these paragraphs contain PCL and to identify which PCL categories these text spans belong to. For this step, we used the BRAT rapid annotation tool \cite{stenetorp2012brat}\footnote{https://brat.nlplab.org/}. 
Note that each paragraph might contain one or more text spans with PCL, which may be assigned to the same or to different categories.
Table \ref{tab:kw-category} shows how many spans have been labelled with each of the categories.

In Task 2, we compute the IAA for each category, reaching the following agreements: \emph{Unbalanced power relations}: 58.43\%; \emph{Authority voice}: 48.34\%; \emph{Shallow solution}: 56.50\%; \emph{Presupposition}: 52.94\%; \emph{Compassion}: 66.40\%; \emph{Metaphor}: 52.72\%, and \emph{The poorer, the merrier}: 66.72\%. When computing the agreement for the three higher-level categories, we obtain a IAA of 63.02\% for \emph{The Saviour} (\emph{Unbalanced power relations} and \emph{Shallow solution}), 57.21\% for \emph{The Expert} (\emph{Presupposition} and \emph{Authority voice}), and 66.99\% for \emph{The Poet} (\emph{Compassion}, \emph{Metaphor} and \emph{The poorer, the merrier}).

\section{Experiments}

We experiment with a number of different methods to provide baselines for further research in modeling PCL. We consider two settings: predicting the presence of PCL, viewed as a binary classification task (Task 1), and predicting PCL categories, viewed as a multi-label classification task (Task 2).
We evaluate the following methods:
\begin{itemize}
\item \textbf{SVM-WV.} We use paragraphs embeddings as the input for a Support Vector Machine implemented with SciKit-Learn. To create the paragraphs embeddings, we use the average of the standard 300 dimensional Word2Vec Skip-gram word embeddings trained on the Google News corpus \cite{mikolov2013distributed}. For Task 1, the parameters that were selected after hyper-parameter tuning were C=10, gamma=`scale', kernel=`poly', while for Task 2 we found that C=100, gamma=`scale', kernel=`rbf' yielded the best results on the validation data. 
\item \textbf{SVM-BoW.} We use a TF-IDF weighted Bag-of-Words representation of the paragraphs as input to an SVM, also implemented with SciKit-Learn. In this case, the hyperparameters that were selected are C=10, gamma= `scale', kernel= `rbf' for Task 1 and C=100, gamme=`scale', kernel= `linear' for Task 2.
\item \textbf{BiLSTM.} We used a bidirectional LSTM, using the same Word2Vec embeddings as SVM-WV to represent the individual words. As hyper-parameters, we used 20 units for each LSTM layer and a dropout rate of 0.25\% at both the LSTM and classification layers. We trained for 300 epochs, using the Adam optimizer, with early stopping and a patience of 10 epochs. 
\item \textbf {Fine-tuned Language Models.} We fine-tune a BERT language model \cite{devlin2018bert} for sequence classification. We considered two variants of this method, were we respectively used the BERT-large-cased and BERT-base-cased pre-trained models.
To further explore the performance of language models, we also fine-tuned a RoBERTa-base \cite{liu2019roberta} model, which can be viewed as an optimized version of BERT, and a DistilBERT \cite{sanh2019distilbert} model, which is a lighter and faster variant of BERT. In all cases, we trained the model for 10 epochs with a batch size of 32. For reproducibility, we fixed the random seeds at 1 in all cases.
%
\item \textbf{Random.} To put the results in context, we include a classifier that relies on random guessing, choosing the positive class with 50\% probability in Task 1, and independently selecting each label with a probability of 50\% in Task 2.
\end{itemize}

For both Task 1 and Task 2 we used 10-fold cross validation for all the experiments. For the BiLSTM models, we used 10\% of the training data in each fold as a validation set for early stopping. For the SVM models, we instead tuned the hyper-parameters using Grid Search Cross-Validation.  
As mentioned before, for Task 1 we view paragraphs labelled with 0 or 1 as negative examples, and the remaining paragraphs, labelled with 2, 3 or 4, as positive examples. The results are reported in terms of the precision, recall and F1 score of the positive class. 
Task 2 is viewed as a paragraph-level multi-label classification problem, where each paragraph is assigned a subset of the PCL category labels. Therefore, in these baselines, span boundaries are not used as part of the training data. 
We report the precision, recall and F1 score of each of the individual category labels.

\begin{table}[!t]
\centering
\footnotesize
\begin{tabular}{lccc}
\toprule
                    & \multicolumn{1}{c}{\textbf{P}} & \multicolumn{1}{c}{\textbf{R}} & \multicolumn{1}{c}{\textbf{F1}} \\ \midrule
\textbf{SVM-WV}     & 46.53                          & 57.80                          & 47.10                           \\
\textbf{SVM-BoW}    & 49.95                          & 40.48                          & 40.59                           \\
\textbf{BiLSTM}     & 62.61                          & 54.43                          & 57.75                                 \\
\textbf{Random}     & 24.82                          & 50.65                          & 33.31                           \\
\textbf{BERT-base}  & 72.39                          & 63.27                          & 67.33                           \\
\textbf{RoBERTa}    & 73.08                          & 68.51                          & \textbf{70.63}                           \\
\textbf{DistilBERT}  & 70.73                          & 66.17                          & 68.24                           \\
\textbf{BERT-large} & 57.06                          & 51.50                          & 53.91                           \\ \bottomrule
\end{tabular}%

\caption{Results for the problem of detecting PCL, viewed as a binary classification problem (Task 1).}
\label{tab:ResultsTask1}
\end{table}

\begin{table}[!h]
\centering
\resizebox{\textwidth}{!}{%
\begin{tabular}{lcccrcccrcccrrrr}
\hline
                          & \multicolumn{3}{c}{\textbf{SVM-WV}}                                    &                      & \multicolumn{3}{c}{\textbf{SVM-BoW}}   & \textbf{} & \multicolumn{3}{c}{\textbf{BiLSTM}}                                   & \textbf{} & \multicolumn{3}{c}{\textbf{Random}}                                    \\
                          & P                         & R                         & F1             &                      & P           & R           & F1         &           & P                         & R                         & F1            &           & \multicolumn{1}{c}{P} & \multicolumn{1}{c}{R} & \multicolumn{1}{c}{F1} \\
\textbf{Unb. power rel.}  & 82.51                     & 85.37                     & 83.82          & \multicolumn{1}{c}{} & 80.02       & 78.91       & 79.21      &           & \multicolumn{1}{r}{83.94} & \multicolumn{1}{r}{84.58} & 83.92         &           & 71.63                 & 49.89                 & 58.54                  \\
\textbf{Authority voice.}    & 40.85                     & 37.27                     & 37.96          & \multicolumn{1}{c}{} & 33.01       & 37.94       & 34.83      &           & \multicolumn{1}{r}{42.54} & \multicolumn{1}{r}{21.27} & 25.71         &           & 21.33                 & 42.01                 & 28.07                  \\
\textbf{Shallow solu.}   & 57.86                     & 50.72                     & 53.49          & \multicolumn{1}{c}{} & 43.20       & 39.05       & 40.39      &           & \multicolumn{1}{r}{64.06} & \multicolumn{1}{r}{31.84} & 40.46         &           & 22.27                 & 55.53                 & 31.67                  \\
\textbf{Presupposition}  & 46.88                     & 42.87                     & 44.28          & \multicolumn{1}{c}{} & 40.01       & 39.73       & 38.12      &           & \multicolumn{1}{r}{52.44} & \multicolumn{1}{r}{36.02} & 41.15&           & 22.53                 & 49.81                 & 30.86                  \\
\textbf{Compassion}         & 68.31                     & 70.41                     & 69.13          & \multicolumn{1}{c}{} & 62.25       & 62.25       & 60.92      &           & \multicolumn{1}{r}{74.48} & \multicolumn{1}{r}{69.86} & 71.34         &           & 49.03                 & 52.64                 & 50.56                  \\
\textbf{Metaphor}       & 37.93                     & 32.80                     & 34.71          & \multicolumn{1}{c}{} & 29.53       & 29.03       & 28.63      &           & \multicolumn{1}{r}{7.83} & \multicolumn{1}{r}{1.99}  & 3.14         &           & 20.12                 & 48.36                 & 28.29                  \\
\textbf{The p., the mer.} & 40.00                     & 12.17                     & 17.89          & \multicolumn{1}{c}{} & 5.00        & 1.43        & 2.22       &           & \multicolumn{1}{r}{0.00}  & \multicolumn{1}{r}{0.00}  & 0.00          &           & 4.40                  & 55.58                 & 8.08                   \\ \hline
                          & \multicolumn{3}{c}{\textbf{BERT-large}}                                & \textbf{}            & \multicolumn{3}{c}{\textbf{BERT-base}} & \textbf{} & \multicolumn{3}{c}{\textbf{RoBERTa}}                                  & \textbf{} & \multicolumn{3}{c}{\textbf{DistilBERT}}                                \\
                          & P                         & R                         & F1             &                      & P           & R           & F1         &           & P                         & R                         & F1            &           & \multicolumn{1}{c}{P} & \multicolumn{1}{c}{R} & \multicolumn{1}{c}{F1} \\
\textbf{Unb. power rel.}  & \multicolumn{1}{r}{84.28} & \multicolumn{1}{r}{93.35} & 88.55          &                      & 84.47       & 93.53       & 88.70      &           & 85.84                     & 93.34                     & \textbf{89.4} &           & 84.11                 & 92.44                 & 88.01                  \\
\textbf{Authority voice.}    & \multicolumn{1}{r}{54.24} & \multicolumn{1}{r}{52.65} & \textbf{53.06} &                      & 54.52       & 43.60       & 47.43      &           & 56.34                     & 48.00                     & 50.9          &           & 51.73                 & 37.16                 & 41.71                  \\
\textbf{Shallow solu.}   & \multicolumn{1}{r}{70.93} & \multicolumn{1}{r}{52.59} & 59.67          &                      & 71.08       & 49.64       & 57.47      &           & 69.09                     & 55.62                     & \textbf{61.0} &           & 72.80                 & 45.21                 & 54.89                  \\
\textbf{Presupposition}  & \multicolumn{1}{r}{60.42} & \multicolumn{1}{r}{59.71} & 59.61          &                      & 59.94       & 55.92       & 57.22      &           & 60.95                     & 58.90                     & \textbf{59.7} &           & 60.32                 & 49.04                 & 53.60                  \\
\textbf{Compassion}         & \multicolumn{1}{r}{78.56} & \multicolumn{1}{r}{76.66} & 77.46          &                      & 77.85       & 76.29       & 76.92      &           & 78.83                     & 77.67                     & \textbf{78.1} &           & 74.17                 & 74.80                 & 74.37                  \\
\textbf{Metaphor}       & \multicolumn{1}{r}{58.51} & \multicolumn{1}{r}{31.48} & 40.09          &                      & 62.81       & 27.93       & 38.21      &           & 59.36                     & 35.74                     & \textbf{43.4} &           & 65.15                 & 27.60                 & 37.93                  \\
\textbf{The p., the mer.} & \multicolumn{1}{r}{23.33} & \multicolumn{1}{r}{8.50}  & 11.90          &                      & 0.00        & 0.00        & 0.00       &           & 40.83                     & 15.00                     & \textbf{20.5} &           & 0.00                  & 0.00                  & 0.00                   \\ \hline
\end{tabular}%
}
\caption{Results for the problem of categorizing PCL, viewed as a paragraph-level multi-label classification problem (Task 2).}
\label{tab:categories_classification}
\end{table}


The results of Task 1 are summarized in Table \ref{tab:ResultsTask1}. As can be seen, all of the considered methods clearly outperform the random baseline. Unsurprisingly, the BERT-based methods achieve the best results, with RoBERTa performing slightly better than DistilBERT and BERT-base. The performance of BERT-large is surprisingly weak compared with the other BERT-based models, performing worse than the BiLSTM. This suggests that BERT-large is more prone to over-fitting, given the relatively small number of training examples. 
Table \ref{tab:categories_classification} shows the results we obtained in Task 2. RoBERTa outperforms the rest of the models in all the categories except for Authority voice, where BERT-large gets the best results. We can also notice the fairly good performance of the SVM methods. In some categories, such as \emph{Methaphors}, the SVM-WV model performs almost on par with DistilBERT and BERT-base and outperforms the BiLSTM results. For \emph{The poorer, the merrier} it outperforms all the models except for RoBERTa. 

Comparing the results for different categories, we can see that \emph{Unbalanced power relations} appear relatively easy to detect. This is not unexpected, given that the presence of words such as \emph{us}, \emph{they}, \emph{must} or \emph{help} are strong and common indicators of such language. For similar reasons, instances of \emph{Compassion} appear relatively easy to detect. \emph{The poorer, the merrier} is the least represented category in the entire dataset, with just 64 samples, which can explain the poor results for this category. However, the poor performance for the \emph{Metaphor} category cannot be explained in this way, given that the number of training examples for this category is higher than the number of examples for \emph{Shallow solution} and very similar to the number of examples for \emph{Authority voice}. More generally, while some of the differences in performance are due to variations in the number of training examples, the categories with the weakest performance also tend to be those that require some forms of world knowledge. For instance, to detect presuppositions, we need to determine whether the assumption which is made is reasonable or not. Similarly, detecting shallow solutions requires assessing the quality of the proposed solution, which can clearly be challenging.

To get further insights into the dataset, Table \ref{tab:wrong_pred} shows some examples of paragraphs from Task 1, their gold labels and the predictions by RoBERTa. There are three correctly classified instances and seven misclassified examples (i.e.\ three false negatives and four false positives). In many cases, we can see words and phrases that are often used in PCL, but which are not actually used in a condescending context, causing the model to predict false positives. For instance, in the seventh example, excess of adjectives and flowery wording, e.g.\ \emph{shocking failures} and \emph{excruciating pressure}, are often used in PCL fragments from the \emph{Compassion} category. In this example, however, it is used in a political context, without being condescending towards any particular group. In the fifth example, the model misclassifies the paragraph as not contaning PCL. In this case, we have an example of the category \emph{The poorer, the merrier}, which all models struggle to detect. Surprisingly, this category has the highest inter-annotator agreement in the annotation of the dataset. This suggest that, while for human annotators it is very easy to identify cases of this category, the models struggle to detect such cases. In Table \ref{tab:misclassified_cats}, some incorrect predictions from Task 2 are presented. Among others, these examples illustrate how RoBERTa struggles to distinguish between presuppositions and authority voices, which are often incorrectly predicted together. Shallow solutions are also often neglected by RoBERTa. A particularly clear case is the last example, where recognizing the presuppositions and shallow solutions in the text will require external knowledge of the situation and the needs of those affected. We can also see examples where the occurrence of a particular structure of language appears to mislead RoBERTa, e.g. \emph{to open the doors wider for [...]}, in the fourth example, seems to lead the model to bet for a shallow solution. \emph{Methaphors}, as in this same example, are also difficult to identify for RoBERTa in this context. 
\begin{table}[!t]
\centering
\footnotesize
\begin{tabular}{rp{370pt}r}
\toprule
\multicolumn{1}{l}{\textbf{Pred.}} & \multicolumn{1}{c}{\textbf{Paragraph}}&
\multicolumn{1}{l}{\textbf{Gold}} \\ \midrule
pos.                                                                                       & After Vatican controversy, McDonald's helps feed homeless in Rome.                                                                                                                                                                                  & pos.                               \\
\midrule
pos.                                                                                       & From his personal story and real-life encounters with poor families, manpower correspondent Toh Yong Chuan suggested shifting the focus from poor parents who repeatedly make bad decisions to their children (Lifting families Out of poverty: Focus on the children; last Thursday).                                                                                                                                           & pos.                               \\
\midrule
pos.                                                                                        & He said their efforts should not stop only at creating many graduates but also extended to students from poor Families so that they could break away from the cycle of poverty.                                                                                                                                                                                                                                                                             & pos.                              \\
\midrule
neg.                                                                                       & ``The biggest challenge is the no work policy. I think that refugees who come here, or asylum seekers, they're unable to work and they have kids here - their kids are stateless. That's really the cause of a lot of stress in the community.''                                                                                                                                 & pos.                               \\
\midrule
neg.                                                                                       & ``The people of Khyber Pakhtunkhwa are resilient. I did not see hopelessness on any face,'' he said.                                                                                                                                                                                                                                                                                                          & pos.                               \\
\midrule
neg.                                                                                       & Teach kids to give back: When Kang runs summer camps with kids, she includes ``Contribution Fridays'' - the kids work together as a team to make sandwiches for the homeless and dole out the food in shelters. & pos.                                     \\
\midrule
pos.                                                                                       & These shocking failures will continue to happen unless the Government tackles the heart of the problem - the chronic underfunding of social care which is piling excruciating pressure on the NHS, leaving vulnerable patients without a lifeline.                                                                 & neg.                              \\ 
\midrule
pos.                                                                                       & Lilly-Hue: His ability to make sure our family is never in need - his sacrificial self.                                                                 & neg.                              \\ 
\midrule
pos.                                                                                       & Any Kenyan small-scale farmer with such an income could not be said to be hopelessly mired in agrarian destitution. But of course, nothing in life is ever so simple as to allow for neat and precise answers.                                                                 & neg.                              \\ 
\midrule
pos.                                                                                       & Selective kindness: In Europe, some refugees are more equal than others.                                                           & neg.                              \\
\bottomrule
\end{tabular}%
\caption{Examples of incorrect predictions made by RoBERTa in Task 1.}
\label{tab:wrong_pred}
\end{table}

\begin{table}[!h]
\centering
\footnotesize
\begin{tabular}{p{350pt}|p{30pt}|p{30pt}}
\toprule
\multicolumn{1}{c}{\textbf{Paragraph}}  & \multicolumn{1}{|l|}{\textbf{Gold}} & \multicolumn{1}{l}{\textbf{Pred}} \\ 
\midrule

[...] The blacks want all our farmland without compensation. Give it to them. Let the farmers flock into the cities and make a new life for themselves . With their resilience I am sure it will not be so difficult for them to establish a new, happy and productive life. They will have no money but the clothes on their back to start off with , but that is what so many immigrant Americans had to face. Through guts, determination and sheer will power, they rose above it all, and look what America is today. & unba, pres, comp, merr      & unba, auth, pres, comp, meta \\               
\midrule

According to the foundation , a number of children between the ages of six and 14 homeless and roaming the streets is becoming alarming. & comp      & unba, comp        \\ 
\midrule
The photo of a Hyderabad traffic policeman feeding an elderly homeless woman has gone viral , earning him accolades from social media users [...]. & unba, shal   & unba  \\ 
\midrule
Practical ways to open the doors wider for our disabled  & unba, meta      & unba, shal                  \\ 
\midrule
He could have also taken his condition to mean he must be disabled from seeking to live for others. He could have degenerated into self pity as many do, wallowing in the muddy fields of self-obsession and low self esteem. Yusuf did not; everything was not about his immediate interests, but a social impact that touched even the lives of strangers [...]. & unba, comp, meta, merr      & auth, pres, comp \\ 
\midrule
She called on the general public to volunteer to donate blood and that way rescue the lives of patients in need of blood transfusion. & unba, auth  & unba, auth, meta\\ 
\midrule
For now the families are staying with friends and family. During the day they clean up the debris left by the fire, hoping that someone will come to their rescue. They received emergency relief packs, but they are still in need of clothes, beds, blankets and kitchen appliances. & unba, shal, pres, comp      & unba, comp \\ 
\bottomrule

\end{tabular}

\caption{Examples of incorrect predictions made by RoBERTa in Task 2.}
\label{tab:misclassified_cats}
\end{table}

\section{Conclusions and Future Work}
We have introduced the Don't Patronize Me! dataset, which is aimed at introducing the NLP community to the 
challenge of identifying and categorizing Patronizing and Condescending Language (PCL) towards vulnerable communities. 
As another contribution of this paper, we also introduced a two-level taxonomy of PCL categories, which was used for annotating the dataset. 
Our exploratory analysis shows that identifying condescending or patronizing texts is a difficult challenge, both for human judges and for NLP systems. Apart from the subtle and subjective nature of PCL, a particular challenge comes from the fact that accurately modelling such language often requires knowledge of the world and common sense (e.g.\ to assess whether a proposed solution is shallow, or whether a particular presupposition is warranted). Nonetheless, we found that both identifying PCL (Task 1) and categorizing occurrences of PCL (Task 2) is feasible, in the sense that non-trivial results can be achieved, with BERT-based approaches outperforming simpler methods.
Future work will include the development of new models for both detecting and categorizing PCL. In addition, we plan to continue to extend the Don't Patronize Me! dataset with more paragraphs from news stories, as well as text fragments from different sources, such as social media or NGO campaigns, to create a useful and updated resource for the community.

\section*{Acknowledgements}

The annotation of the Don't Patronize Me! dataset has been funded by a Kaggle Open Data Research Grant and by the Data Innovation Research Institute (DIRI) at Cardiff University.
We would like to thank especially the two main annotators of this project, Mireia Roig Mirapeix and Edurne Morillo Garcia.


\bibliographystyle{coling}
\bibliography{coling2020}

\end{document}